\documentclass[lettersize,journal]{IEEEtran}
\usepackage{amsmath,amsfonts}
\usepackage{algorithmic}
\usepackage{algorithm}
\usepackage{array}
\usepackage{subfig}
\usepackage{textcomp}
\usepackage{stfloats}
\usepackage{url}
\usepackage{verbatim}
\usepackage{graphicx}
\usepackage{cite}
\begin{document}

\title{LightFuse: Lightweight CNN based Dual-exposure Fusion}

\author{Ziyi~Liu\thanks{Ziyi Liu, Svetlana Yanushkevich, and Orly Yadid-Pecht are with the Schulich School of Engineering, University of Calgary, Calgary T2N 1N4, Canada (e-mail: ziyi.liu1@ucalgary.ca; syanshk@ucalgary.ca; orly.yadid-pecht@ucalgary.ca).}, Jie~Yang\thanks{Jie Yang is with the School of Engineering, Westlake University, Hangzhou 310024, China. Corresponding
author e-mail: yangjie@westlake.edu.cn}, Svetlana Yanushkevich, Orly Yadid-Pecht,
\IEEEmembership{Fellow,~IEEE} %
\thanks{This work was supported in part by the Alberta Innovates Technology Futures under Grant RT735246, and in part by the Natural Sciences and Engineering Research Council of Canada under Grant RT731684 and Grant 10023039.
}
}



\maketitle

\begin{abstract}
Deep convolutional neural networks (DCNNs) have aided high dynamic range (HDR) imaging recently and have received a lot of attention. The quality of DCNN-generated HDR images has overperformed the traditional counterparts. However, DCNNs are prone to be computationally intensive and power-hungry, and hence cannot be implemented on various embedded computing platforms with limited power and hardware resources. Embedded systems have a huge market, and utilizing DCNNs' powerful functionality into them will further reduce human intervention. To address the challenge, we propose $LightFuse$, a lightweight CNN-based algorithm for extreme dual-exposure image fusion, which achieves better functionality than a conventional DCNN and can be deployed in embedded systems. Two sub-networks are utilized: a $GlobalNet$ ($G$) and a $DetailNet$ ($D$). The goal of $G$ is to learn the global illumination information on the spatial dimension, whereas $D$ aims to enhance local details on the channel dimension. Both $G$ and $D$ are based solely on depthwise convolution ($D\_Conv$) and pointwise convolution ($P\_Conv$) to reduce required parameters and computations. Experimental results show that this proposed technique could generate HDR images in extremely exposed regions with sufficient details to be legible. Our model outperforms other state-of-the-art approaches in peak signal-to-noise ratio (PSNR) score by 0.9 to 8.7 while achieving 16.7 to 306.2 times parameter reduction\footnote{The code is made publicly available at https://github.com/Taichi-Pink/LightFuse-Lightweight-CNN-based-Dual-exposure-Fusion}. 
%(有没有关于图像质量的描述？).
%（实现了XXX的PSNR，SSIM 等等,XXX的功耗等等，要有量化指标）
\end{abstract}


\section{Introduction}
\IEEEPARstart{D}{ynamic} range is defined as the ratio of the intensity of the brightest point to that of the darkest point in a scene or an image. The dynamic range of digital images is mainly determined by the saturation current of the photoelectric response and dark current of the imaging device. Also, the dynamic range of each device is fixed and usually narrower than that of the natural scene, leading to overexposed or underexposed images that miss details in the brightest or the darkest regions. Hence, high dynamic range (HDR) imaging technology is used to overcome this deficiency in modern cameras. It can show rich appearances of natural scenes such as lighting, contrast, and details and are extremely important in many applications such as photography and autonomous vehicles \cite{paul2018application}. Some professional hardware devices have been devised to capture HDR \cite{tocci2011versatile, nayar2000high}. However, these devices are either expensive or bulky. Researchers are resolving this issue on the algorithmic level and have designed plethoric algorithms to reproduce a greater range of luminosity. In particular, the most popular algorithm to generate an HDR image is known as multi-exposure fusion (MEF) \cite{mertens2007exposure}, which takes a sequential low dynamic range (LDR) image in the same scene under different exposure levels with a normal camera and incorporates these images into one that contains more details and information than any of the input images.

A considerable number of MEF algorithms have been proposed over the years, which can generally be classified into two main categories: traditional and deep learning-based (DL). Traditional MEF methods \cite{6423909, ma2017robust, ma2017multi} require manually designed mathematical transformation and fusion equations and apply the functions on an image bracket for blending multiple exposures of the same scene into a single image. However, it is hard to manually design appropriate equations to pick the important features from each input image. DL approaches are more likely to avoid this problem because trained neural networks (NNs) can spontaneously establish complex relationships between source images and fused images through training on a volume of sample data. For example, Wu et al. \cite{wu2018deep} designed the first non-flow-based deep MEF framework with a model size of 64MB. Kalantari et al. \cite{kalantari2017deep} used a convolutional neural network (CNN) as their learning model and presented three different MEF architectures, demanding 4.08 MB of memory. Among these methods, the classical CNN as a preferred network structure has extravagant power consumption, excessive parameters, and heavy computational resources occupation issues, making them hard to be deployed on embedded computing applications. 

Commonly, CNN methods deployed on mobile devices are actually implemented on the cloud to access sufficient computing resources \cite{wang2018deep}, and the mobile devices are harnessed to receive and present processed results from the cloud. However, cloud computing risks the safety of user privacy and introduces data transceiving delays that are critical to real-time applications such as driverless cars \cite{liu2019edge}, security surveillance \cite{nikouei2018smart}, and industrial automation \cite{li2018deep}. Implementation of DL models on local devices can avoid delays and privacy vulnerabilities. Nevertheless, this brings new challenges to designing lightweight models for power- and resource-constrained devices. To address these difficulties, this paper presents $LightFuse$ architecture. To our best knowledge, this is the first lightweight MEF method. It consists of two networks, $GlobalNet$ ($G$) and $DetailNet$ ($D$), which find the global- and local-related patterns separately. With pointwise convolutions ($P\_Conv$) and depthwise convolutions ($D\_Conv$), our model prevents high-computational costs and eludes underfitting by mimicking the process procedures of standard convolutions.

The organization of this paper is as follows. Related work is introduced in section 2. We describe our proposed dual-exposure fusion algorithm in section 3 and compare it with the state-of-the-art methods in section 4. We conclude the chapter in section 5.

\section{Related Work}
\subsection{Depthwise Separable Convolution}
\label{sub:flops_introduction}
\begin{figure}[!t]
\centering
\subfloat[Standard Convolution]{\includegraphics[width=\linewidth]{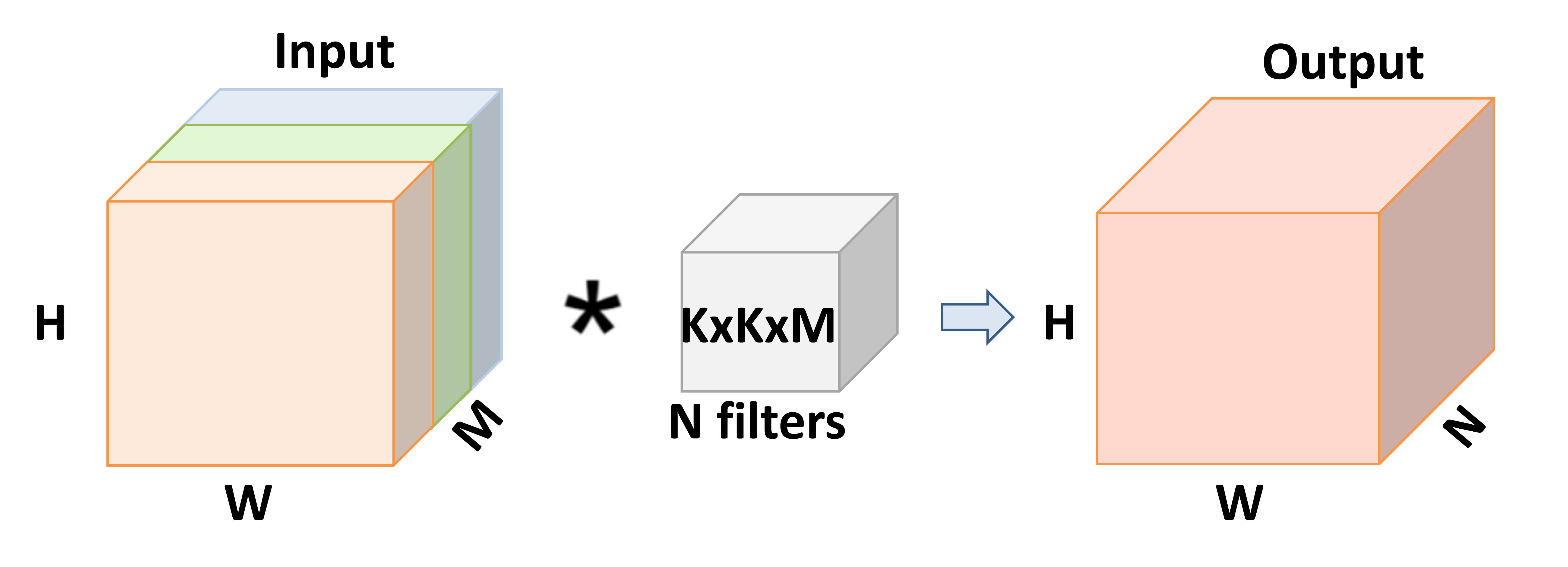}
\label{fig:fig_first_case}}
\hfil
\subfloat[Depthwise Separable Convolution]{\includegraphics[width=\linewidth]{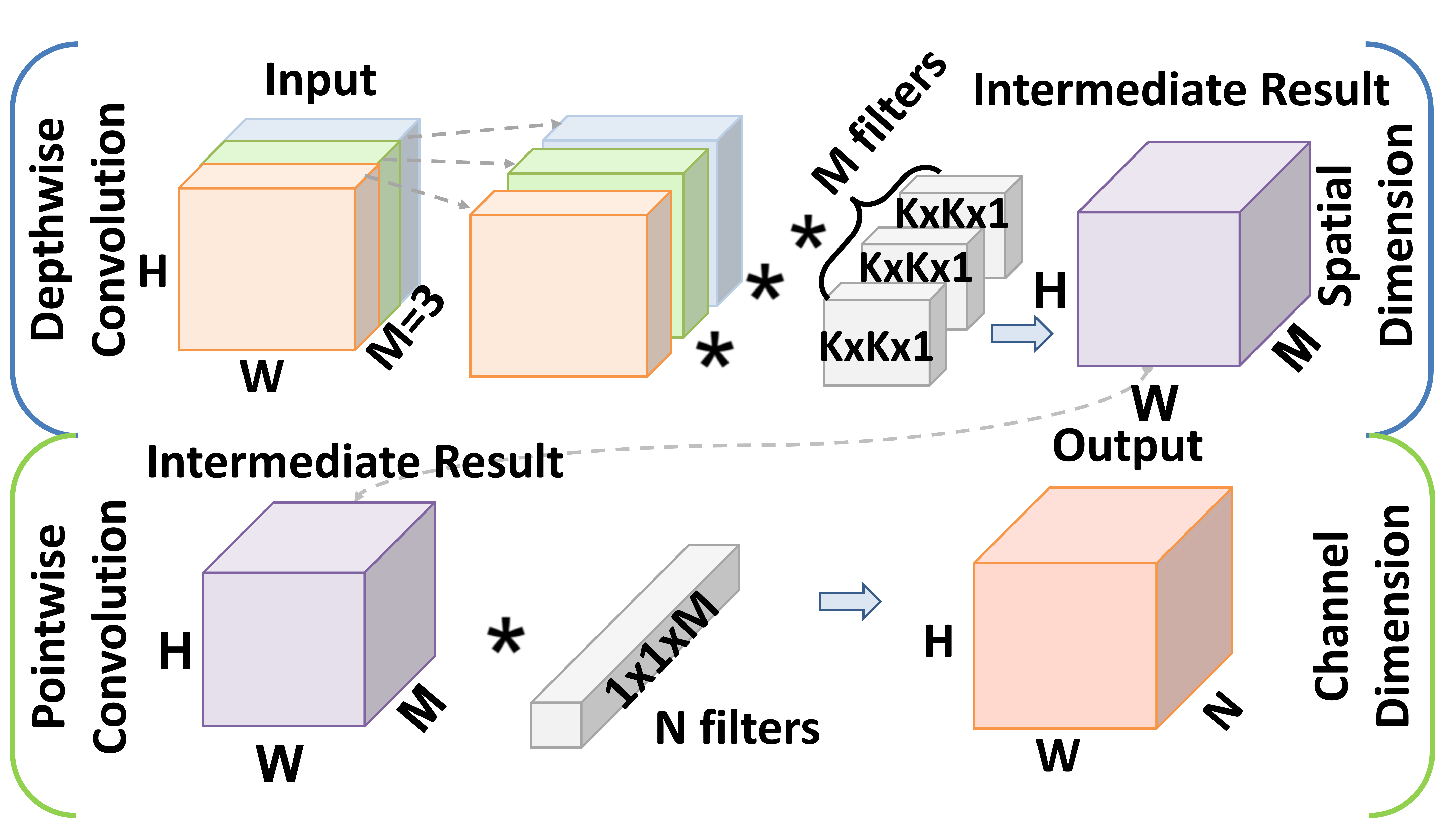}
\label{fig:fig_second_case}}
\caption{The standard convolution in (a) is replaced by two stages: depthwise convolution and pointwise convolution in (b) to build a depthwise separable convolution.}
\label{fig:1}
\end{figure}

The standard convolution in Figure \ref{fig:fig_first_case} tries to learn features from three-dimensional data, dealing with both spatial dimensions and a depth (channel) dimension. The purpose of creating a depthwise separable convolution ($DS\_ Conv$) is to decompose the standard convolution into two stages to improve efficiency to make the operation easier. More precisely, the computational cost is significantly reduced by replacing a conventional convolution with a $DS\_Conv$ composed of two subtasks, a depthwise convolution ($D\_Conv$) and a pointwise convolution ($P\_Conv$). As depicted in Figure \ref{fig:fig_second_case}, a $D\_Conv$ does spatial convolutional calculations with $K \times K$ kernels first, in which two-dimensional kernels are convoluted with an input data channel by channel; then, the $P\_Conv$ that is equivalent to a $1 \times 1$ standard convolution is applied to integrate information for all channels.

If the size of an input feature map is $H \times W \times M$ and that of a produced feature map is $H \times W \times N$, a $K \times K$ standard convolution requires $K \times K \times M \times N \times W \times H$ floating-point operations (FLOPs). But the $DS\_Conv$ solely needs $K \times K \times M \times W \times H$ FLOPs to get the intermediate result and another $M \times N \times W \times H$ FLOPs to get the final output, which is $\frac{1}{N} + \frac{1}{K^2}$ of the standard convolution's operations. In addition, convolutional layers carry the most parameters of NNs that are stored in the main memory. The standard convolution layer has $K \times K \times M$ weights per filter. This layer has $N$ filters, so the total parameters are $K \times K \times M \times N$ plus an additional $N$ bias value. In the same way, a $D\_Conv$ requires $K \times K \times M + N$ parameters, and the quantity of parameters a $P\_Conv$ costs is $M \times N + N$. Since a $DS\_Conv$ only has one bias operation, its parameter quantities are $K \times K \times M + M \times N + N$. Normally, the number of weights is much larger than that of biases, so we neglect biases to easily demonstrate that $DS\_Conv$ is $\frac{1}{N} + \frac{1}{K^2}$ of the standard convolution's parameters. In brief, the channel value $N$ is the dominant factor for parameters and computational cost as $K^2 \ll N$. In most research \cite{krizhevsky2012imagenet, simonyan2014very, szegedy2015going, he2016deep}, an exponential of two ($2^5-2^{12}$) is preferred as the number of channels. As the depth increases, the number of channels for each layer is doubled, leading to exponential growth as well as a high computational requirement. In this situation, we may consider the $DS\_Conv$ instead of the conventional convolution to avoid large computing and memory budgets.

\subsection{Lightweight Neural Network}
Many techniques have been developed for compression and quantization of complex NN models and have demonstrated encouraging results \cite{hinton2015distilling, han2015deep, wu2016quantized}. However, it is more important to mitigate computational costs in the initial phase of model design to reduce training time and complexity of NN compression later on. Developing a tiny NN with comparable accuracy has recently attracted people’s attention. Chollet et al. \cite{chollet2017xception} propose a model named Xception, in which the performance is superior to Inception V3 \cite{szegedy2016rethinking} by replacing the Inception module with $DS\_ Convs$. Similarly, Howard et al. \cite{howard2017mobilenets} build a MobileNet based on $DS\_ Convs$. To further enhance the performance of mobile models on various tasks, a new mobile network, MobileNetV2, is described \cite{sandler2018mobilenetv2}, containing a novel inverted residual architecture with a linear bottleneck. Moreover, Zhang et al. \cite{zhang2018shufflenet} introduce pointwise group convolution and channel shuffle operations, and compose them together as a ShuffleNet unit to create a computation-efficient model that surpasses the MobileNet. To conclude, small NNs can achieve the same or even better performance than deeper NNs. In this thesis, we introduce a lightweight network architecture for multi-exposure image fusion that achieves real-time fusion performance without surrendering accuracy compared with other recent systems \cite{mertens2009exposure, ram2017deepfuse, ma2019deep, xu2020mef, zhang2020rethinking}.

\begin{table}
\fontsize{10}{12}\selectfont
\centering
    \begin{tabular}{|l|c|c|}
    \hline
    Input Size & Operator & Output Size\\ \hline 
    $W \times H \times 6$ & 3 x 3 Separable Conv & $W \times H \times 32$\\ \hline
    $W \times H \times 32$ & 3 x 3 Separable Conv & $W \times H \times 32$\\ \hline
    $W \times H \times 32$ & 3 x 3 Separable Conv & $W \times H \times 3$\\ 
   \hline
    \end{tabular}
    \caption{$T\_CNN$ architecture based on depthwise separable convolutions.}
\label{tab:first_CNN}
\end{table}

\begin{figure*}[tb!]
\centering
\includegraphics[width=\linewidth]{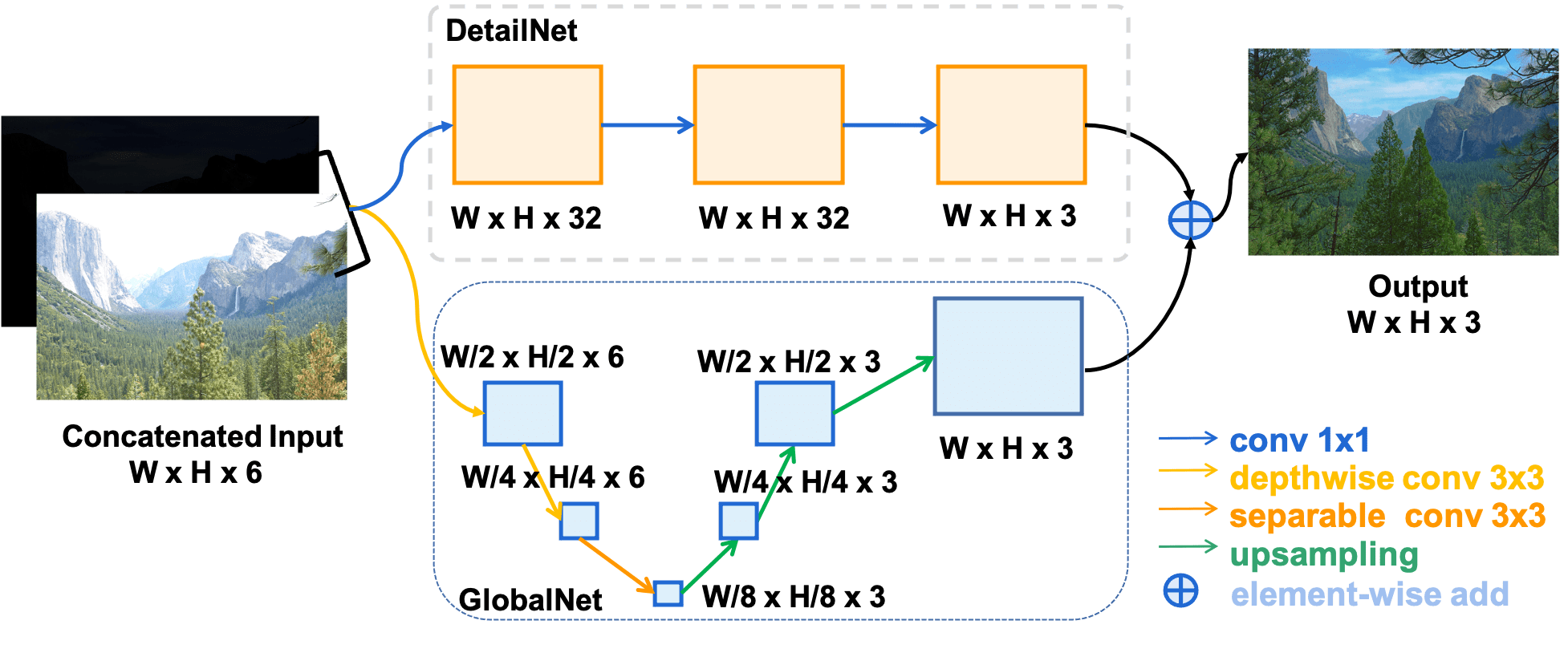}
\caption{An overview of our lightweight image fusion method. Our proposed method comprises two networks: a $GlobalNet$ and a $DetailNet$. Given extreme overexposed and underexposed images as inputs, the $GlobalNet$ extracts the global illumination information, while the $DetailNet$ enhances the local details. Eventually, the output image is obtained by adding the features obtained from the two sub-nets, followed by a Tanh operation.}
\label{fig:overall_structure}
\end{figure*}

\section{Approach}
\label{lightfuse1}
The architecture of $LightFuse$ is depicted in Figure \ref{fig:overall_structure}. It takes extreme overexposed and underexposed image pairs as inputs. A $GlobalNet$ is utilized to extract the global structural information of the input images on the spatial dimension, and a $DetailNet$ is adopted to enhance local details on the channel dimension.

\subsection{Trial Convolutional Neural Network}
Convolution is widely used in computer vision tasks. As a CNN model goes deeper, the amounts of computations and parameters of convolutional layers increase significantly. Therefore, variants of standard convolutions start rising to reduce complicated computation but still retain competitive visual effects. 

In this study, we mainly focus on $P\_Conv$ and $D\_Conv$. The $P\_Conv$ is essentially a $1 \times 1$ standard convolution ($1 \times 1 \; Conv$) that has a small receptive field \cite{araujo2019computing} but combines local details from all other feature representations; the kernel size of a $D\_Conv$ empirically is $3 \times 3$ or $5 \times 5$ for a sufficient receptive field that is big enough to capture the spatial relationships among local contour fragments. Thus, we characterize the fused image with information from these two paths: the local details and the global shapes, which is reflected in the network architecture.

\begin{figure*}[tb!]
    \centering
       \includegraphics[width=\linewidth]{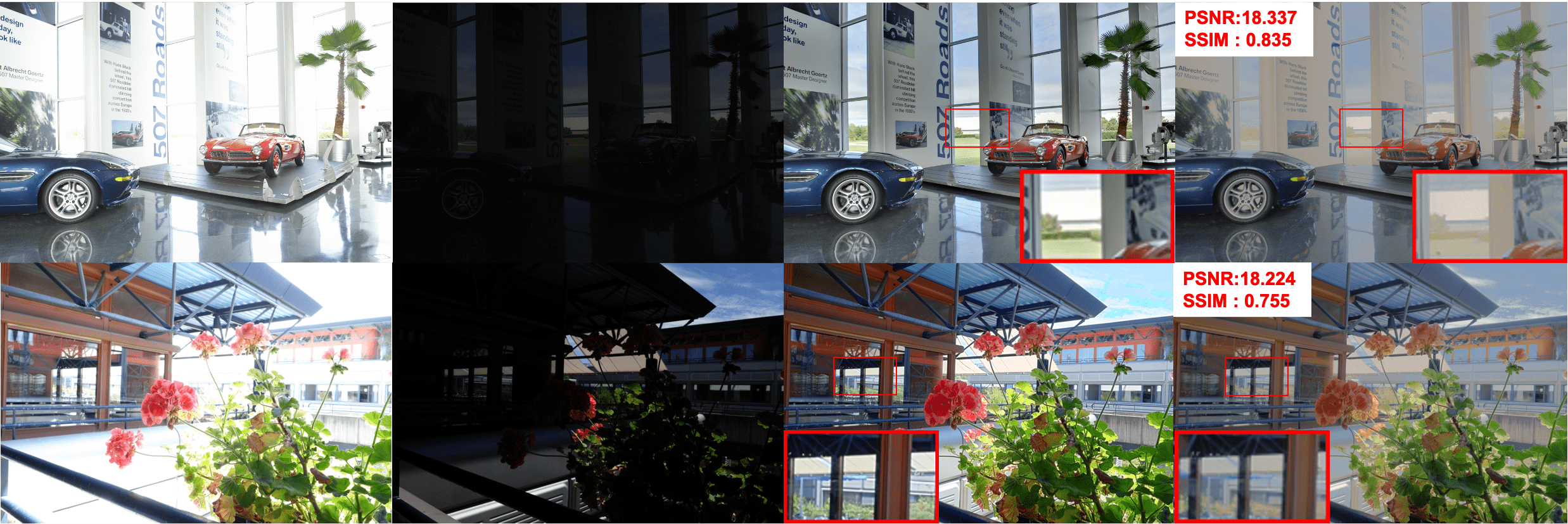}
       \hfill
    \caption{Obtained HDRs through $T\_CNN$ architecture. From left to right: over-exposed and under-exposed source images, reference images, and results of $T\_CNN$. Metric values (mentioned in Section \ref{metrics}) are drawn in the upper left corner.}
\label{fig:t_cnn}
\end{figure*}

Table \ref{tab:first_CNN} illustrates a trial CNN ($T\_CNN$) architecture, consisting of three depthwise separable convolution ($DS\_Conv$, as mentioned in Section \ref{sub:flops_introduction}) layers. From Figure \ref{fig:t_cnn}, we observe that the $DS\_Conv$ structure has the potential to extract and reconstruct desirable information from differently exposed source images. In the second scene, clouds are missing in the overexposure but clear in the underexposure. $T\_CNN$ can extract cloud information from the overexposure and maintain it in the output even when the ground-truth (see Figure \ref{fig:t_cnn}) has artifacts. Then, we enhance $T\_CNN$ by creating and training two separate networks (Figure \ref{fig:overall_structure}), one that focuses on the channel dimension and the other that operates on the spatial dimension. By separating the channel and spatial dimensions into separate networks, we avoid the original proposed $T\_CNN$ existing problem that alternatively trains on the spatial and channel dimensions, resulting in unstable performance.

\subsection{Network Architecture}
\subsubsection{Framework Overview}
Most MEF algorithms \cite{mertens2009exposure,li2012fast,wu2018deep,niu2021hdr} expect a sequential LDR image to generate high-quality HDR images, hence they require many LDR images to capture the whole dynamic range of a scene, causing large storage requirements, processing time, and power budget. However, the exposure differences between successive LDR images in an exposure stack are slight, resulting in information redundancy. Following \cite{ram2017deepfuse, xu2020mef, zhang2020rethinking}, we utilize two extremely exposed images with complementary information; together, they contain all the information necessary for image fusion.

The architecture of $LightFuse$ is illustrated in Figure \ref{fig:overall_structure}. It consists of a $GlobalNet$ and a $DetailNet$. The $G$ has a U-Net-shaped \cite{ronneberger2015u} structure, made up of an encoder (left side) and a decoder (right side). $D$ consists of three $P\_Conv$ layers, each followed by a ReLU nonlinearity. The number of channels in $D$ is increased from six to 32 through the first layer to expand the channel space, and then decreased to three through the third layers to combine the channel-related information. We employ $1 \times 1$ kernels in these layers to maintain detailed information and ensure that no crucial information is missed. Moreover, the last layer of $LightFuse$ is an addition layer that merges the learned global and local information from two sub-nets, followed by a Tanh activation function. Feature map addition has its unique advantages and is ubiquitously used in the visual field \cite{he2016deep, ram2017deepfuse}. The addition saves more calculation and storage compared to multiplication and concatenation operations. In \cite{he2016deep}, He et al. propose a residual learning framework that adds the input feature maps of the residual module with its outputs, to increase the network depth without the degradation of accuracy. We choose an addition operation for the similar purpose that increases the speed of the prediction process without crumbling the performance.

\subsubsection{GlobalNet Architecture}
 In U-Net, skip-connections are used between the encoder and decoder to prevent information loss with the increase in depth of the architecture. For $LightFuse$, we did not add these connections, because the depth of $G$ is low. Besides, they involve adding the intermediate results from the encoder to the current decoder layers, which requires more data storage and extra memory access, and is unsuitable for resource-limited platforms. Additionally, the encoder in $G$ consists of three convolutional layers, performing subsequent down-sampling operations on their respective inputs. For the down-sampling, the number of channels of the first two $D\_Conv$ layers (closest to the input layer) with strides of two stays the same as six, until the third $DS\_Conv$ layer sets it as three in correspondence with the final outputs. The decoder contains three up-sampling layers, operating nearest-neighbor interpolations to reconstruct the compressed representations. In brief, $G$ filters the input channels to capture the global information but does not combine them to create new features. Thus, we introduce the $D$ to concentrate on the channel dimension by projecting the input channel onto a large channel space and learning new features from this high-dimensional space. 
 
\subsubsection{DetailNet Architecture}
We implement a fusion architecture \cite{xiao2017exploring} in $D$, in which only an input tile of the first layer is required to obtain one element in the final output layer. All the necessary intermediate tiles can be computed without storing and retrieving the intermediate data to and from off-chip memory. Thus, only the final layer's output feature maps are stored in off-chip memory, which reduces the pressure of off-chip memory bandwidth.

\begin{figure}[tb!]
\centering
\includegraphics[width=\linewidth]{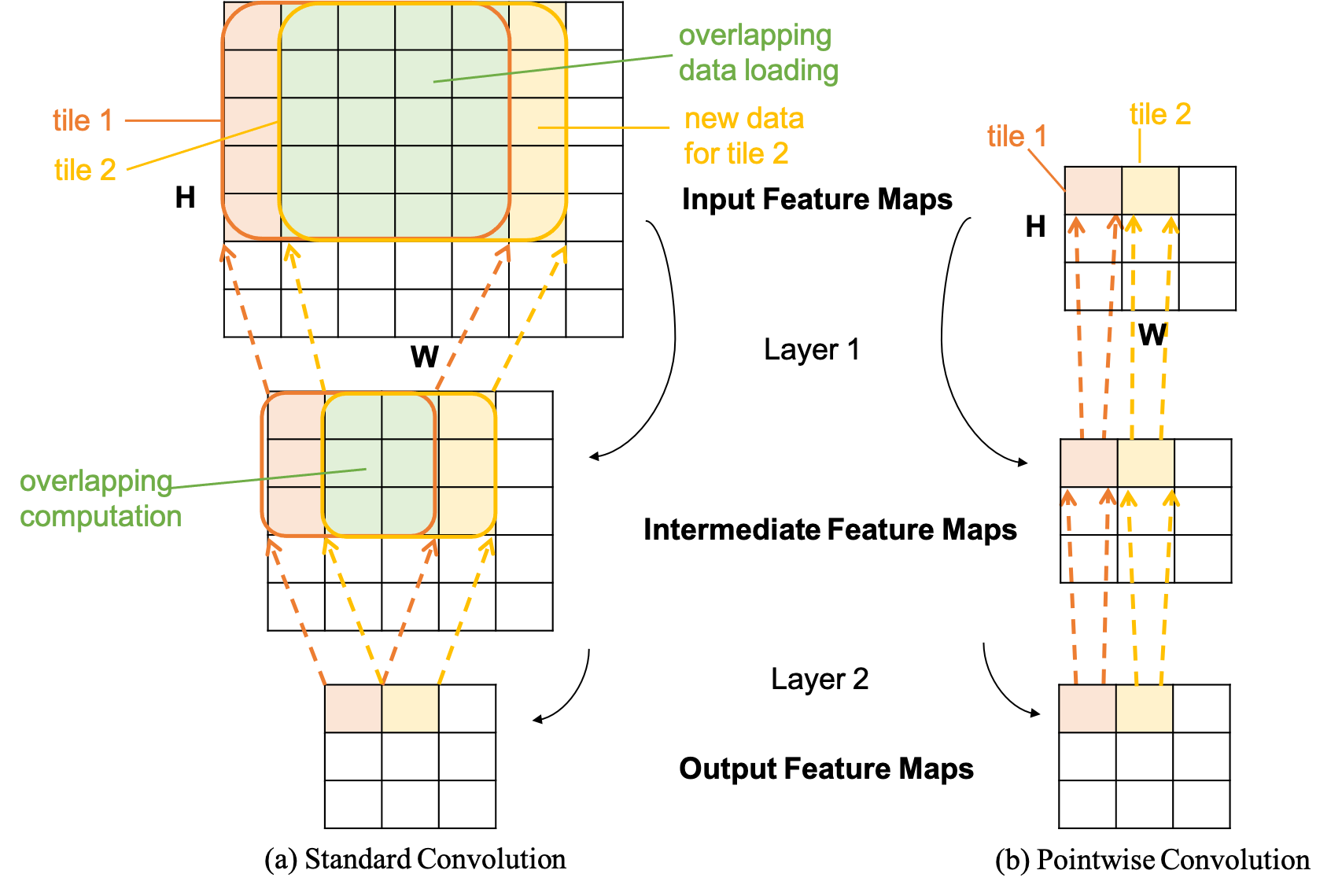}
\caption{A fusion example of two layers with a stride of one. The kernel size of (a) is $3 \times 3$, and that of (b) is $1 \times 1$.}
\label{fig:fpga_data_reuse_P}
\end{figure}

Figure \ref{fig:fpga_data_reuse_P} displays a fusion example in which two layers are calculated simultaneously, and the intermediate results are cached on the chip. This figure also compares the data efficiency between a $3 \times 3$ standard convolution and a $P\_Conv$ that works essentially as a $1 \times 1 \; Conv$. Each element in the output feature maps of Layer 2 in the $3 \times 3$ standard convolution depends on a $5 \times 5$ input tile. As the stride is one, this standard convolution loads overlapping data and repeatedly computes these data, which is portrayed in Figure \ref{fig:fpga_data_reuse_P}a.
Conversely, each output element in $P\_Conv$ relies on a $1 \times 1$ input tile of Layer 1 without considering any spatial elements around the input element, so the $P\_Conv$ loads sequential data in accordance with the current operational order. As a result, fused $P\_Conv$ layers are able to reduce off-chip bandwidth requirements and on-chip computational resources. Compared with the $3 \times 3$ standard convolution, $P\_Conv$ is more appropriate to be applied in a fusion architecture.

\begin{figure}[tb!]
    \centering
	\includegraphics[width=0.9\linewidth]{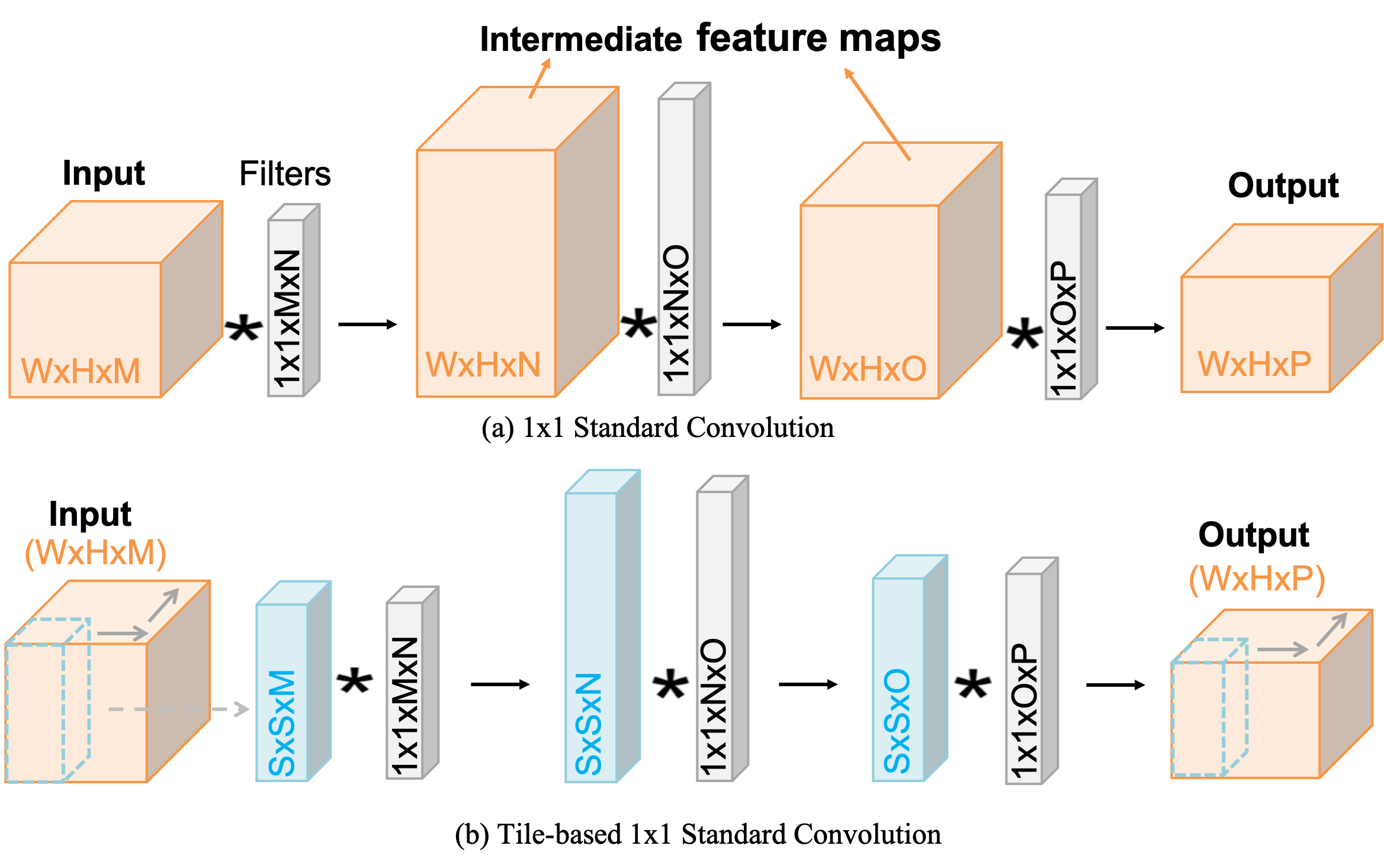} 
    \caption{Implementation comparison of a normal $1 \times 1 \; Conv$ and a tile-based $1 \times 1 \; Conv$.}
  \label{fig:1x1conv2d_block} 
\end{figure}

The three $P\_Conv$ layers in $D$ are fused. Figure \ref{fig:1x1conv2d_block}b illustrates the process: we partition the input feature maps into tiles of size $S \times S \times M$ along channel depth with no overlap; each tile is convoluted end-to-end sequentially and concatenated together in the last layer (the third $P\_Conv$ layer). Here, $S \times S$ is the spatial size of a tile, and it can be adjusted corresponding to the specifications of resource-constrained devices to satisfy memory limitations. This process will not increase the number of weights and biases because all the operations in one layer use the same weight and bias. In other words, one layer solely has one weight and bias like normal $1 \times 1 \; Conv$. Besides, a tile-based $1 \times 1 \; Conv$ has the same function as the $1 \times 1 \; Conv$ module in Keras \cite{chollet2015} but less data read and write to off-chip. This results from the fused layer eliminating the intermediate feature maps store and fetch as displayed in Figure \ref{fig:1x1conv2d_block}.

\subsection{Loss Function}
To train the network, a combination of two different loss functions are used: perceptual loss \cite{johnson2016perceptual} and mean square error loss (MSE). The perceptual loss compares high-level abstractions between the generated image and its ground-truth image by passing these two through a pre-trained network that applies convolution layers to extract features and patterns, and calculates the distance between these two images.
\begin{equation}
L_{perceptual} = \sum_{i} \parallel f_{i}(I_{out}) - f_{i}(I_{label}) \parallel_{1}
\label{equ:vgg}
\end{equation}

In our work, VGG19 \cite{simonyan2014very} is chosen as the feature extractor noted as $f$ in Equation \ref{equ:vgg}, and $I_{out}$ and$I_{label}$ represent the network output and ground truth image. 
\begin{equation}
L_{mse} = \frac{1}{n} \sum_{j=1}^{n} (I_{out_{j}} - I_{label_{j}})^2
\label{equ:mse}
\end{equation}

The MSE loss function is given above. Here, $n$ refers to the number of pixels in $I_{out}$ or $I_{label}$. It is used as a metric for understanding differences between images on a pixel-by-pixel level.

\begin{equation}
L = L_{perceptual} + L_{mse}
\label{equ:loss}
\end{equation}

Our training function is defined in Equation \ref{equ:loss}, a mixture of two different loss functions.

\subsection{Implementation Details}
Adam \cite{kingma2014adam} is an optimization algorithm that iteratively updates neural network models' weights (parameters) based on training data. We use it as the optimizer and set the learning rate and batch size as 0.001 and 20 apart. Although a deeper model may achieve better performance, in this study, we aim to present a simple idea as a reference in the high dynamic range image field. The overall framework can be written in Keras \cite{chollet2015} or other high-level neural network libraries. Contrary to training large models, we use fewer regularization and data augmentation techniques because small models have less trouble with overfitting. The values of training data and labels range from -1 to 1.

%detail custom conv2d and trade-off
\section{Experiments}
\label{lightfuse2}
\subsection{Dataset}
The SICE dataset \cite{cai2018learning} is selected for training and testing the proposed network. This dataset contains 589 scenes, in which the number of images in a multi-exposure image sequences of different scene varies. In order to pair the dual-exposure images, we use the average pixel value of an image to estimate its exposure value (EV). Since the contents in a multi-exposure image sequence are the same, only the exposures are different, we can sort the images according to their exposure values and pick two images with the highest and the lowest EV as our experimental data. The dataset \footnote{https://github.com/csjcai/SICE} contains two parts; in this study, we only use the first part with 360 sequences, because it has more classic pictures the authors obtained from a widely used dataset \cite{nemoto2015visual}. Additionally, we randomly select 80\% of the images for training and use the remaining 20\% for testing. For the sake of carrying out practical training, we crop the original full-size image into $256 \times 256$ patches with a stride of 256, generating about 68,000 patches. During testing, the size of images ranges from $1000 \times 1504$ to $4000 \times 6000$, resulting in an error due to insufficient memory, so we set the input size as $896 \times 1344$ when testing on a Nvidia Tesla V100 GPU.

\subsection{Input Image Pairs}
\begin{figure*}[tb!] 
    \centering
       \includegraphics[width=\linewidth]{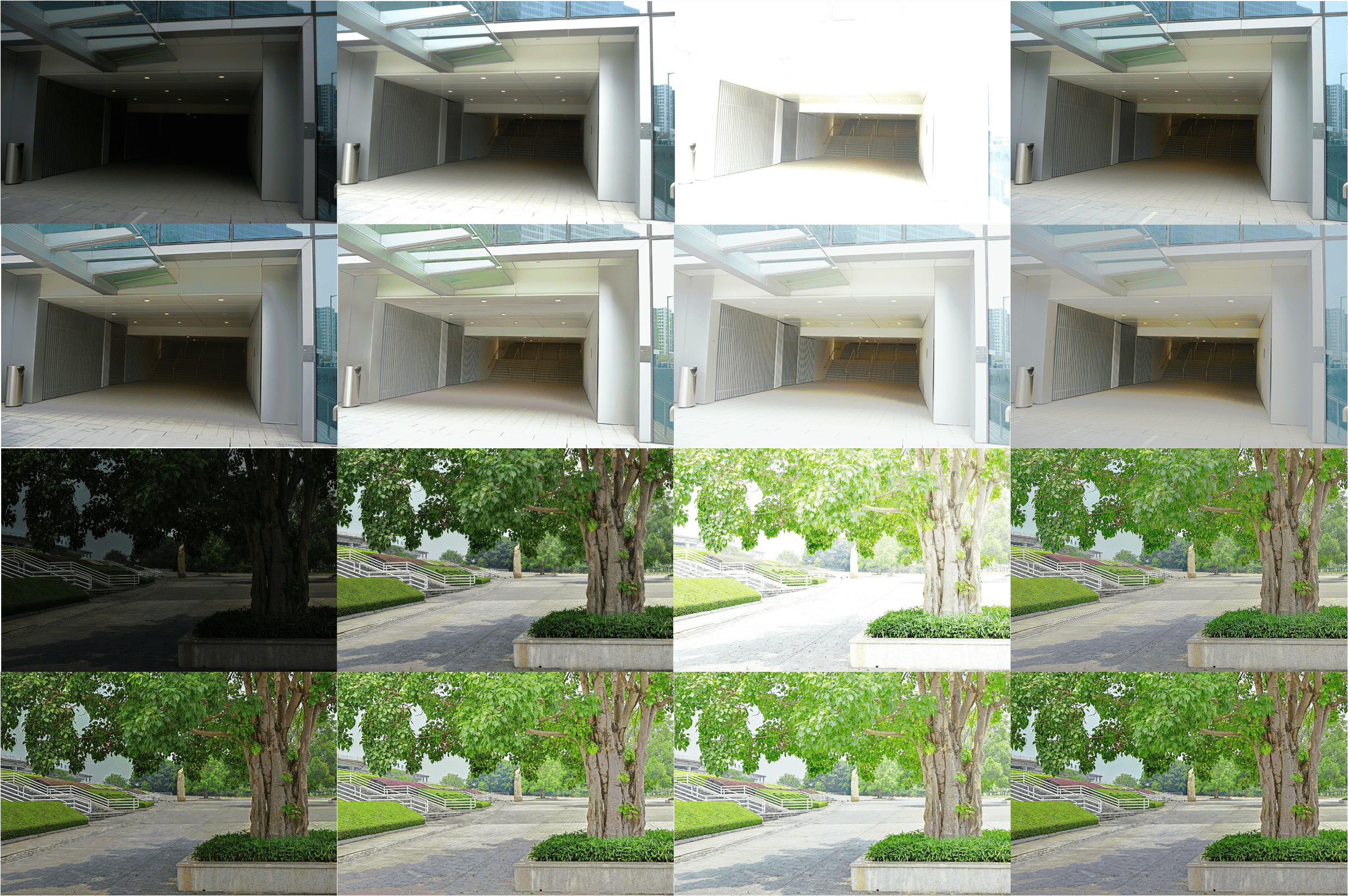}
       \hfill
    \caption{Comparison of image fusion on different image pairs. From left to right: under-, medium-, over-exposed source image, and reference (first row); results of $LightFuse$ on under-medium, medium-medium, medium-over, and under-over image pairs (second row).}
\label{fig:image_pairs}
\end{figure*}

The SICE dataset \cite{cai2018learning} is selected in this work. It does not contain a file listing EV gaps between images, so we regard the average pixel value of an image as the EV. We sort images in each sequence by EV and select input image pairs. In each scene, we seek to fuse the extreme image pair; thus, the vital feature of our model is the capability to extract and fuse complementary information from image pairs. As Figure \ref{fig:image_pairs} shows, we examine this capability on different image pairs, and the results successfully contain contents from both images. Taking the first scene as an example, glasses from the medium exposure and stairs from the overexposure are combined into the medium-over output; on the medium-medium inputs, the image brightness and color saturation are improved in the output; on the under-medium inputs, the information from the two are merged into the produced image.

\begin{figure*}[tb!]
    \centering
       \includegraphics[width=\linewidth]{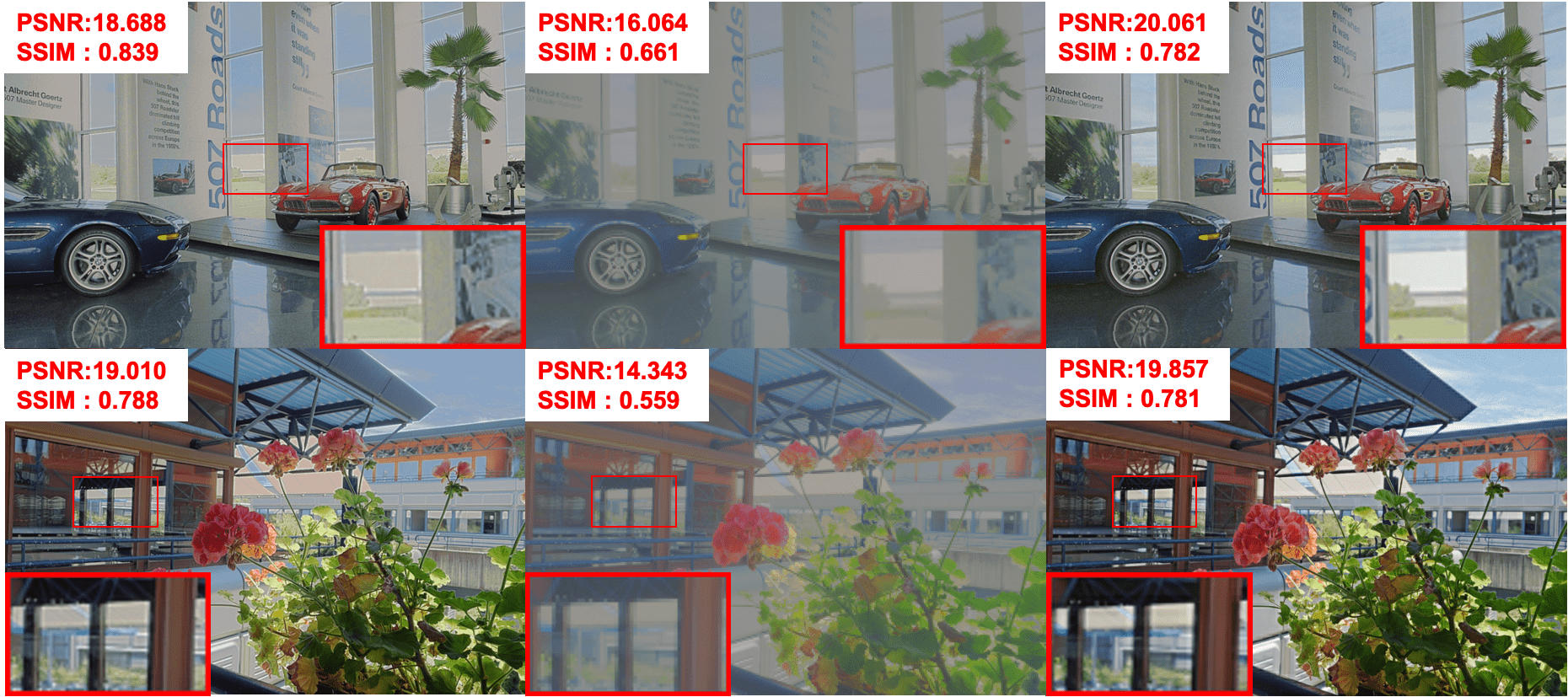}
       \hfill
    \caption{Comparison of fused images from $D$, $G$, and $LightFuse$. From left to right: results of $D$, $G$, and $LightFuse$.}
\label{fig:ablation}
\end{figure*}
\subsection{Ablation Studies}
We conduct ablation studies to interpret the function of each sub-network. We train $G$ and $D$ solely, and then check their performance on the testing sets. As depicted in Figure \ref{fig:ablation}, both $D$ and $G$ incorporate complementary information from source images and revise the improper representations in the references (see Figure \ref{fig:t_cnn}), such as artifacts in the sky area of the second scene. However, the results of $G$ are blurred and merely show contours of objects, e.g., outside views of the labeled window in the first scene are indistinct. On the contrary, the results of $D$ contain more detailed features, but their outlines of the labeled pillars and windows in the second scene are vague. To explain further, the distinct functions of $G$ and $D$ are mainly caused by their differences in channels and kernels. $G$ with larger kernels receives broader receptive fields \cite{luo2017understanding} and captures spatial relationships among local contour fragments; $D$ with $1 \times 1 \; Convs$ expands channel space and extracts fine-grained information from all other feature maps to describe the local details. Combining these two, our model could learn from different scales and receptive fields.

We also observe that results of $D$ (Figure \ref{fig:ablation}) and $T\_CNN$ (Figure \ref{fig:t_cnn}) have close visual effects, but the former has higher metric values. This may be explained by $D$ being composed of three $P\_Conv$ layers only train on the channel dimension and thus can intensively learn the detailed information. Nevertheless, $D$ is not as good as the $LightFuse$ that incorporates $P\_Conv$ and $D\_Conv$ layers. Likewise, $G$ only consists of $D\_Conv$ layers and does not perform well either. Therefore, $D\_Conv$ and $P\_Conv$ are two essential paths to characterize the input images and unite them to obtain great fused results (third column in Figure \ref{fig:ablation}), which is also reflected in the network architecture (Figure \ref{fig:overall_structure}).

\subsection{Performance on Embedded Computing Platforms}
\begin{figure*}[tb!]
    \centering
       \includegraphics[width=\linewidth]{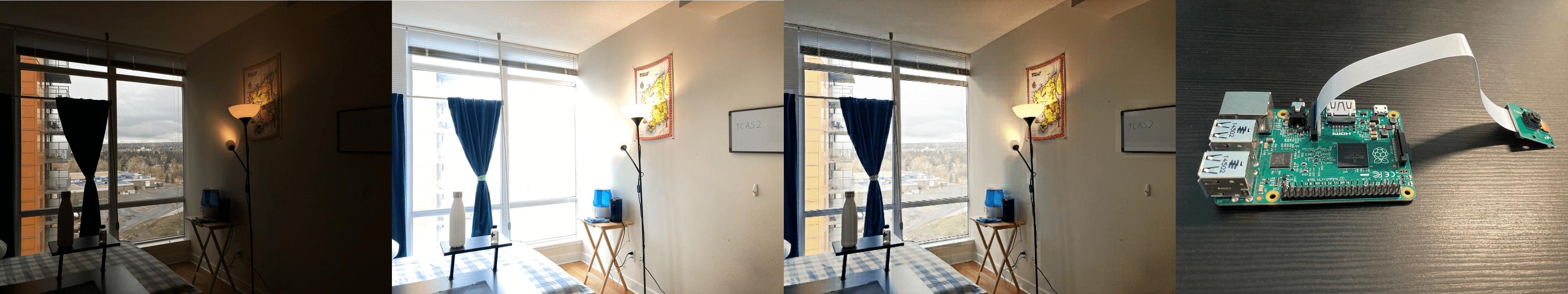}
       \hfill
    \caption{Performance on the Raspberry Pi. From left to right: under-exposed and over-exposed images, fused result, and our system.}
\label{fig:raspberry}
\end{figure*}

To illustrate the performance of our method on embedded systems, we first build a system using a Raspberry Pi 2 model B with 1 GB of RAM and a 900MHz quad-core ARM Cortex-A7 processor, and a 5MP 1080p HD Camera Module. Our model is transferred to the TensorFlow Lite framework and deployed on the board for inference. As illustrated in Figure \ref{fig:raspberry}, in the system, we take two extremely exposed images from a scene in everyday life, and then they are fused to generate the resulting image, which contains the view out of the window and the whiteboard contents.

To further examine the performance of $LightFuse$, we also run it on an FPGA. The target FPGA platform of the system in this study is an Avnet ZedBoard that is used widely by many researchers in academia. This board has a Xilinx Zynq Z-7020 Zynq device with a dual-core ARM Cortex-A9 processor and a traditional FPGA. The ZedBoard also includes a 512MB DDR3, a 4GB SD card, and a 256Mb Quad-SPI flash memory. Its FPGA fabric has 85k logic cells with 280 18kb Block RAM, 220 DSP slices, 53k lookup tables (LUTs), and 106k flip-flops (FFs). 

\begin{figure*}[tb!]
    \centering
       \includegraphics[width=\linewidth]{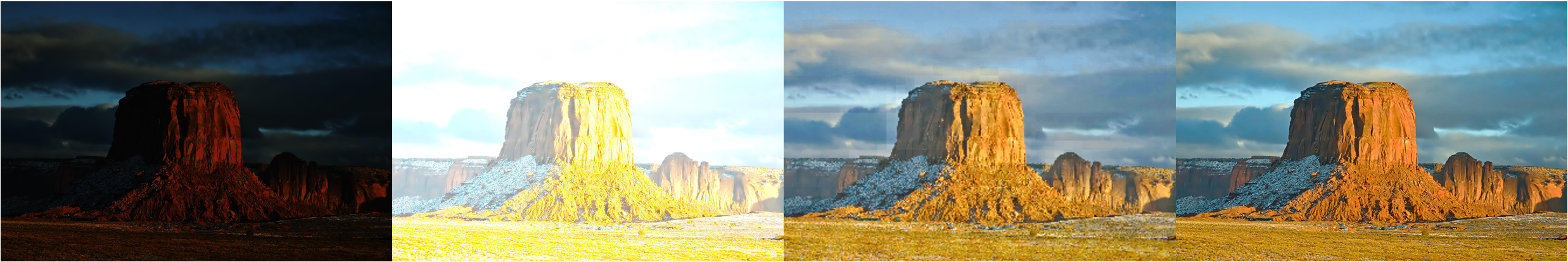}
       \hfill
    \caption{Performance on the FPGA. From left to right: under-exposed and over-exposed source images, fused result, and reference.}
\label{fig:fpga_result}
\end{figure*}

We implement the above-proposed image fusion architecture into the FPGA platform for real-time HDR image generating (see appendix for the system block diagram of the FPGA accelerator). During the experimental test, the accelerator achieves 30 frames per second, which can compete with human eye frames per second. As well, the on-chip resource utilization of ZedBoard is reported in Table \ref{tab:zedboard_resources}. Among them, BRAM is the block RAM with a size of 18KB in the FPGA, which cache parameters of convolutional layers and temporary results. DSP48E is a programmable arithmetic computing unit that supports the parallel computing of multiplication and addition of weights and feature maps. We can tell that our $LightFuse$ accelerator has almost fully utilized the hardware resource on Zedboard. Besides, Figure \ref{fig:fpga_result} demonstrates the generated result, which is visually pleasing with a PSNR score of 22.3.

\begin{table}
\fontsize{10}{12}\selectfont
\centering
    \begin{tabular}{|l|c|c|c|c|}
    \hline
    Resource & BRAM & DSP48E & FF & LUT\\ \hline
    Used & 52 & 211 & 38502 & 48468 \\ \hline
    Available & 280 & 220 & 106400 & 53200\\ \hline 
    Utilization (\%) & 18.6 & 95.9 & 36.2 & 91.1\\ \hline
    \end{tabular}
    \caption{Hardware Resources Consumption of ZedBoard}
\label{tab:zedboard_resources}
\end{table}

\subsection{Evaluation and Comparison}
\subsubsection{Qualitative Comparison}
\begin{figure*}[tb!]
 \centering
\includegraphics[width=\linewidth]{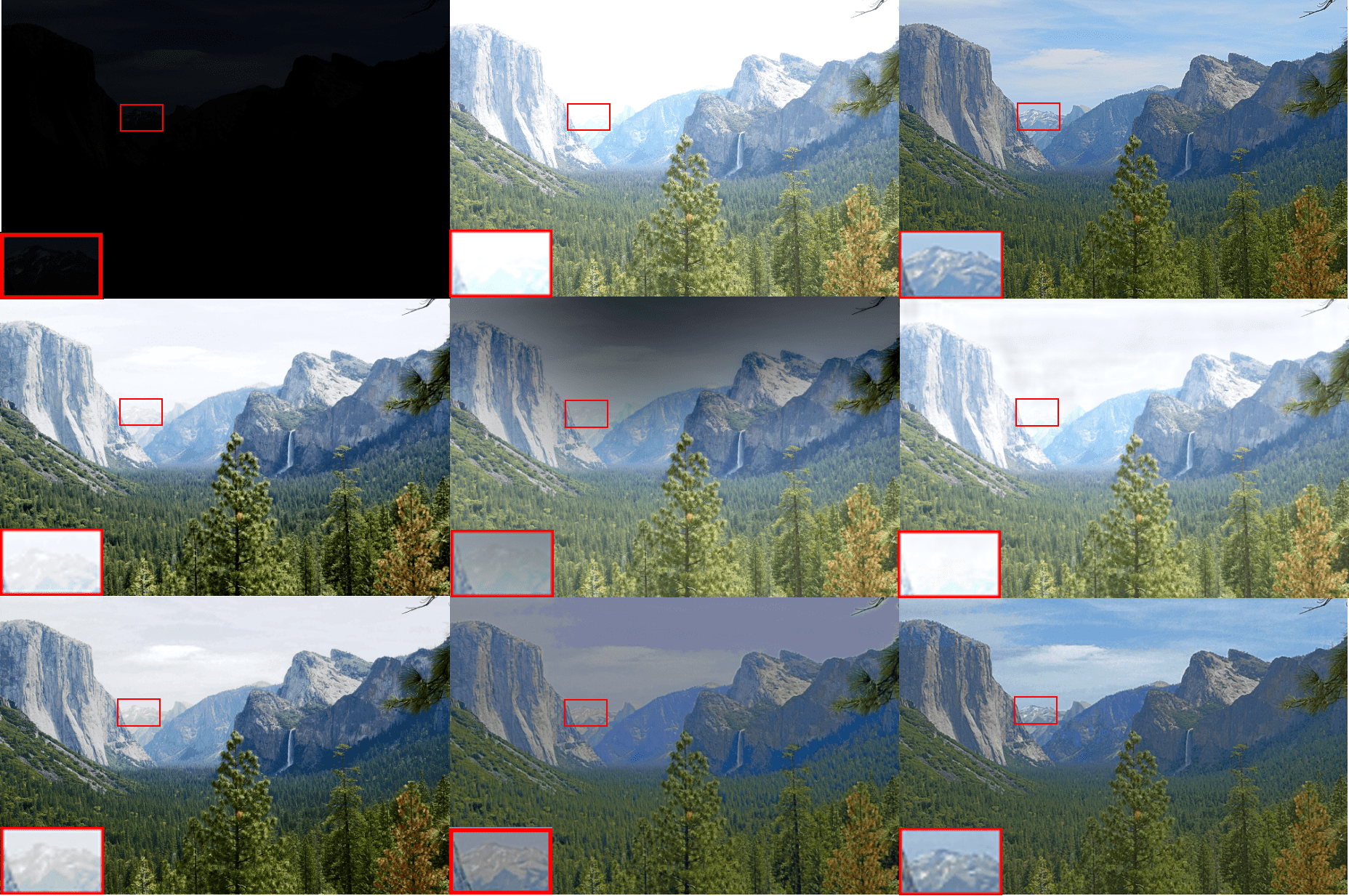}
  \caption{Comparative results of $LightFuse$ with existing methods. From left to right: under-, over-exposed source image, Reference; DeepFuse \cite{ram2017deepfuse}, Mertens09 \cite{mertens2009exposure}, MEF-Net \cite{ma2019deep}; PMGI \cite{zhang2020rethinking}, MEF-GAN \cite{xu2020mef}, $LightFuse$.}
\label{fig:quality} 
\end{figure*}

We compare $LightFuse$ with the previous five MEF methods on unified exposure image pairs from \cite{cai2018learning}. As displayed in Figure \ref{fig:quality}, the result of Mertens09 \cite{mertens2009exposure} tends to be dark, as their pyramid reconstruction cannot guarantee the produced intensities within the range of 0 to 1. In contrast, the results of DeepFuse \cite{ram2017deepfuse} and MEF-Net \cite{ma2019deep} contain too much brightness that leads to over-contrast in the region where the sky meets the mountain. The over-contrast results from their loss function MEF-SSIM \cite{ma2015perceptual}, which is usually employed in unsupervised training and cannot learn mappings from source images to the ground truth directly. In addition, the produced images from MEF-Net and MEF-GAN \cite{xu2020mef} suffer from evident dark spots, especially in the regions with the strongest brightness in the overexposed image. Furthermore, MEF-Net learns weight maps for each input and then sums all the Hadamard products of the inputs with the corresponding weight maps to yield results. These weight maps prefer well-exposed regions; thus, the fused results may not be appealing (i.e., look as good as possible in terms of quality) when both source images miss details in one region. Finally, the output of PMGI \cite{zhang2020rethinking} has clear textures, but its chrominance characteristics are unrealistic, possibly because PMGI is trained on the luminance channel. In conclusion, our operation produces a more visually pleasing scene: the mountain is presented clearly, the sky is bluer than the reference, and the sharpness of the forest is preserved.
\subsubsection{Quantitative Comparison}
\label{metrics}
We list the quantitative comparison results in terms of the image’s peak signal-to-noise ratio (PSNR) and structural similarity index measure (SSIM). PSNR measures the distortion by the ratio of peak value power and noise power. SSIM is well correlated with HVS and designed based on three factors: brightness, contrast, and structure. In Table \ref{tab:computation}, most DL methods have surpassed the traditional one Mertens09 \cite{mertens2009exposure} because the DL models are trained on a preprocessed dataset, where they discover the underlying patterns in images and automatically find the most descriptive and salient features concerning each exposure of a scene. However, the performance of MEF-Net \cite{ma2019deep} is not competitive because it prefers three or more exposure inputs. From Table \ref{tab:computation}, we can see that $LightFuse$ performs the best results on both metrics, compared with other algorithms on the same testing set. In addition, the testing platforms are an Nvidia Tesla V100 GPU and a computer with an Intel Core i7-10870H (2.20GHz) CPU. Mertens09 \cite{mertens2009exposure}, DeepFuse \cite{ram2017deepfuse}, and MEF-Net utilize CPU, while MEF-GAN \cite{xu2020mef} and PMGI \cite{zhang2020rethinking} exploit GPU. We test our model on both platforms.

\subsubsection{Computational Complexity Comparison}
As shown in Table \ref{tab:computation}, we conduct a computational complexity comparison regarding the number of floating-point operations (FLOPs) and parameters. MEF-GAN has a merge block and two paralleled blocks focusing separately on local details and global dependencies. The local module contains five convolutional layers, each with 40 filters. Dense connections are applied on these layers that reuse the features from all the previous layers, increasing the channel number significantly and using many computational resources. Thus, MEF-GAN has the highest computational cost. PMGI is also composed of two paralleled paths: a gradient path that captures the local texture of images and an intensity path that learns image distribution. Each path consists of four convolutional layers with dense connections, and each layer has 16 filters. Compared with MEF-GAN, PMGI does not cost numerous computations. Moreover, the reduced parameters of PMGI did not degrade its performance heavily, as demonstrated in Table \ref{tab:computation}. Consequently, we build our model with three layers in each network. Although our model has fewer parameters, it outperforms PMGI and MEF-GAN due to the particular structures of $P\_Convs$ and $D\_Convs$ that implement the comparable function of standard convolutions.

MEF-Net has four components: a bilinear downsampler, a context aggregation network, a guided filter, and a weighted fusion module. The context aggregation module with seven convolutional layers allocates the most resources. Similarly, DeepFuse, which has five convolutional layers, is divided into three components, including feature extraction layers, a fusion layer for added operations, and reconstruction layers. Its large kernel sizes cause many computations. Conversely, our model structure spends 88.82\% of its computation time in $1 \times 1 \; Convs$, which also has 88.95\% of the parameters, as seen in Table \ref{tab:resource}. 

However, $LightFuse$ enjoys the lowest computational complexity because it restricts most of the computation at a fixed small channel or kernel size, whereas the competing algorithms perform all computation at the standard convolution, which requires a high computational resource. As well, DeepFuse and PMGI are not end-to-end operations. Their well-trained DL models operate on the luminance channel obtained through converting the exposure image stack into YCbCr color space; to get the final RGB results, we need to fuse the output results from the DL models with Cb and Cr channels. For these two methods, we solely consider the time demanded to generate the luminance results in Table \ref{tab:computation}. Moreover, all the algorithms run on the same testing set with the resolution of $896 \times 1344$, to measure their execution time. The computational cost of $LightFuse$ is 3.5x to 6508x faster than others. Notably, our model is even faster than the traditional method Mertens09. Meanwhile, the total number of parameters of $LightFuse$ is around 16.7x and 306.2x less.

\begin{table}
\fontsize{10}{12}\selectfont
\centering
    \begin{tabular}{|l|c|c|}
    \hline
    Type & FLOPs & Parameters\\ \hline  
    1 × 1 Conv & 88.82\% & 88.95\%\\ \hline
    3 × 3 Depthwise Conv  & 10.91\% & 11.05\%\\ \hline
    Up-sampling & 0.27\% & 0.00\%\\ 
   \hline
    \end{tabular}
    \caption{Resource Per Layer Type}
\label{tab:resource}
\end{table}

\begin{table*}
\fontsize{10}{12}\selectfont
\centering
    \begin{tabular}{|l|c|c|c|c|c|}
    \hline
    Algorithm & FLOPs & Parameters & Exe. Time(s) & PSNR & SSIM \\ \hline 
    Mertens09 \cite{mertens2009exposure} & - & - & 0.52 & 14.014 & 0.739  \\ \hline
    DeepFuse \cite{ram2017deepfuse} & $W \times H \times 736,928$ & 369,921& 976.21 & 17.342 & 0.793\\ \hline
    MEF-Net \cite{ma2019deep} & $W \times H \times 52,320$ & 26,305 & 1.92 & 11.495 & 0.668 \\ \hline
    Ours(CPU) & $W \times H \times 2,984$ & 1,574 & 0.15 & 20.224 & 0.797 \\ \hline
    MEF-GAN \cite{xu2020mef} & $W \times H \times 962,976$ & 482,015  & 2.76 & 19.307 & 0.760 \\ \hline
    PMGI \cite{zhang2020rethinking} & $W \times H \times 82,880$ & 41,760 & 0.46 & 18.638 & 0.785 \\ \hline
    Ours(GPU) & $W \times H \times 2,984$ & 1,574 & 0.03 & \textbf{20.224} & \textbf{0.797} \\
   \hline
    \end{tabular}
    \caption{Compare $LightFuse$ with other methods. The cells containing the highest score across all scenes are highlighted in bold.}
\label{tab:computation}
\end{table*}

\section{Conclusion and future work}
We propose a lightweight network for image fusion. Coming from the rationale of depthwise separable convolution, we suggest an exposure fusion framework $LightFuse$ composed of two sub-nets, where the $GlobalNet$ focuses on the global shapes, and the $DetailNet$ seeks to improve local details. This particular architecture reduces model size and latency without sacrificing image quality. Moreover, we implement a fusion architecture in $DetailNet$ to avoid unnecessary data storing and fetching. Our method benefits deploying deep learning architectures on embedded computing platforms, for which we often consider the issue of hardware utilization. Under limited resources, hardware support for any operation requires hardware resources that are preoccupied. Therefore, our method greatly alleviates this problem and achieves a trade-off between hardware flexibility and computing performance. The model was only tested on a Raspberry Pi 2 model B and an Avnet ZedBoard platform. In the near future, the model will be employed on other novel platforms. Also, we will try to resolve other existing issues in the high dynamic range imaging area using lightweight neural networks, such as tone mapping operators and single image enhancers.

\bibliographystyle{IEEEtran}
\bibliography{references}

\appendices
\section{FPGA-based CNN Accelerator for Deep Multi-Exposure Fusion}
\begin{figure*}[tb!]
\centering
\includegraphics[width=\linewidth]{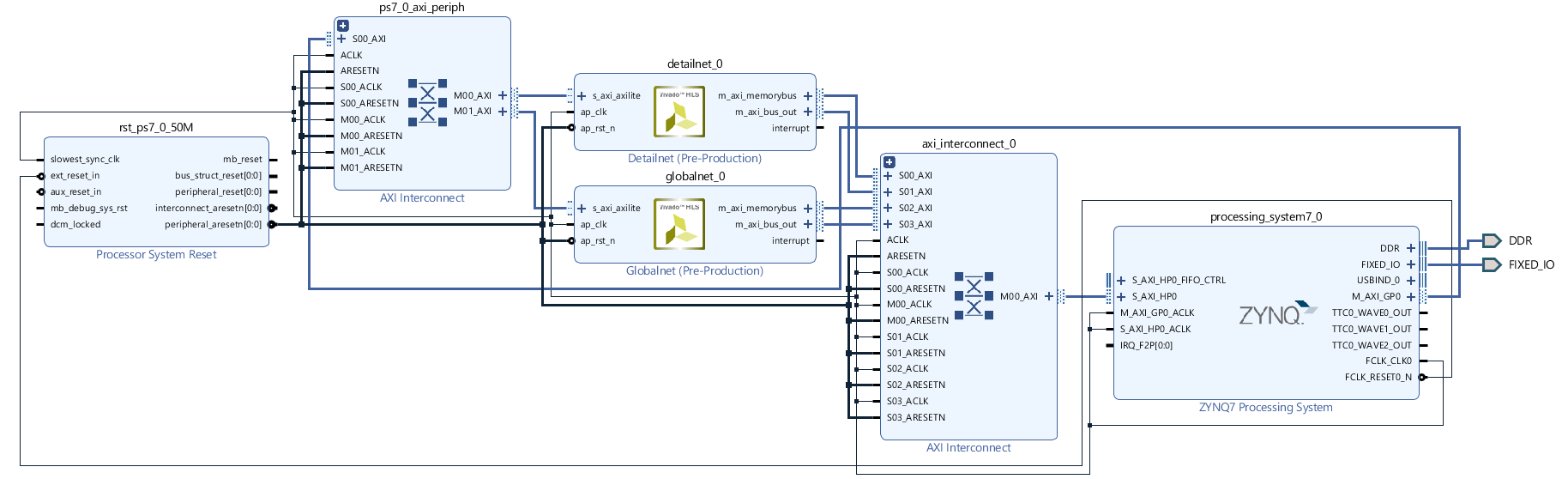}
\caption{System block diagram of the $LightFuse$ FPGA Accelerator.}
\label{fig:system_block_diagram}
\end{figure*}

Figure \ref{fig:system_block_diagram} displays the system block diagram that is generated by the Block Design tool of Vivado Design Suite. It includes a ZYNQ processing system and our designed $DetailNet$ and $GlobalNet$ accelerators.

\end{document}